# Logical Consensus for Distributed and Robust Intrusion Detection

Adriano Fagiolini, *Member, IEEE*, and Antonio Bicchi, *Fellow, IEEE*

*Abstract*—In this paper we introduce a novel consensus mechanism where agents of a network are able to share logical values, or Booleans, representing their local opinions on e.g. the presence of an intruder or of a fire within an indoor environment. We first formulate the logical consensus problem, and then we review relevant results in the literature on cellular automata and convergence of finite–state iteration maps. Under suitable joint conditions on the visibility of agents and their communication capability, we provide an algorithm for generating a logical linear consensus system that is globally stable. The solution is optimal in terms of the number of messages to be exchanged and the time needed to reach a consensus. Moreover, to cope with possible sensor failure, we propose a second design approach that produces robust logical nonlinear consensus systems tolerating a given maximum number of faults. Finally, we show applicability of the agreement mechanism to a case study consisting of a distributed Intrusion Detection System (IDS).

## I. INTRODUCTION

A LARGE number of control problems, especially problems involving a distributed collection of agents, require a form of *consensus* in order to make the system work properly. Consensus problems that are formulated in the control literature generally concern how to reach an agreement on data that can be represented by a real number or vector, such as a temperature, the level of a liquid in a tank, etc. This is achieved by allowing agents to combine local estimates of the quantity of interest, obtained through direct measurement, with the ones shared by the neighbors of a communication graph $G$. A typical form of such *distributed consensus systems* is represented by the following continuous–time linear system (see e.g. [1]–[3]):

$$\dot{x}(t) = A\,x(t) + B\,u(t)\,, \qquad (1)$$

where $A \in \mathbb{R}^{n\times n}$ is a strongly connected doubly–stochastic matrix, $B \in \mathbb{R}^{n\times m}$ is the input matrix, and $u \in \mathbb{R}^m$ is a control law. The flourishing literature on this topic have studied both continuous– or discrete–time, synchronous or asynchronous, and quantized versions of such systems and has provided useful results to determine properties such as characterization of equilibria, and convergence rate [?], [4], [5]. Many control applications such as flocking and formation control may indeed exploit this form of linear consensus.

A. Fagiolini and A. Bicchi are with the Interdepartmental Research Center "E. Piaggio", Faculty of Engineering, Università di Pisa, Italy, {a.fagiolini, bicchi}@centropiaggio.unipi.it.

Nevertheless, the rich literature on *distributed algorithms* shows that many other applications would benefit from availability of *more general forms of consensus*, where participating agents are *de facto* able to reach an agreement on non–scalar quantities. Relevant examples of applications are represented by consensus algorithms studied in Lynch's book [6], or the solution proposed by Cortes *et al.* for achieving consensus on general functions [7]. Other examples are clock synchronization (see e.g. Marzullo's work [8]), and cooperative simultaneous localization and map building [9]. More recently, a distributed IDS for multi–agents has been proposed in [10], where agents cooperate to establish whether a common neighbor acts according to shared logical rules. In all such application scenarios a form of *consensus on intervals or sets* is required and thus solutions based on simple linear consensus can not be applied.

In this perspective, we introduce a novel form of consensus, so–called *logical consensus*, where a number of agents have to decide on the value of a set of *decisions* depending on logical inputs. This problem differs from threshold–based consensus because state variables are Booleans here, and the associated dynamics are qualitatively different. The realization of such logical consensus systems can build upon known results in the literature on cellular automata and convergence of finite–state iteration maps [11], [12]. These results has been recently used for the realization of a content–addressable memory [13]. Our ambitious objective is to develop a *synthesis technique* for logical consensus systems, that can reach a certain level of systematicity as it happens in the linear case. Indeed, under suitable joint conditions on the visibility of agents and their communication capability, we provide an algorithm generating logical linear consensus systems that are globally stable. The solution is optimal in terms of the number of messages to be exchanged and the time needed to reach a consensus. Moreover, to cope with possible sensor failure, we propose a second design algorithm that produces robust logical consensus systems. Detecting and tolerating misbehavior of some agents is an interesting challenge, which has recently received an increasing attention (see e.g. the solution proposed in [14] for a linear consensus system, and the one in [10]).

## II. PROBLEM STATEMENT

We consider control problems requiring computation of a set of $p$ *decisions*, $y_1,\ldots,y_p$, that depend on $m$ logical *events*, $u_1,\ldots,u_m$. Such events may represent e.g. the presence of an intruder or of a fire within an indoor environment. More precisely, for any given combination of input events, we consider a *decision task* that requires computation of the following system of logical functions:

$$\begin{cases} y_1 = f_1(u_1, \ldots, u_m), \\ \quad \vdots \\ y_p = f_p(u_1, \ldots, u_m), \end{cases} \quad (2)$$

where each $f_i : \mathbb{B}^m \to \mathbb{B}$ represents a logical condition on the inputs. Let us denote with $u = (u_1, \ldots, u_m)^T \in \mathbb{B}^m$ the input event vector, and with $y = (y_1, \ldots, y_p)^T \in \mathbb{B}^p$ the output decision vector. We will write $y = f(u)$ as a compact form of Eq. 2, where $f = (f_1, \ldots, f_p)^T$, with $f : \mathbb{B}^m \to \mathbb{B}^p$, is a logical vector function. It is worth noting that computation of $f$ is *centralized* in the sense that it may require knowledge of the entire input vector $u$ to determine the output vector $y$.

Our approach to solve the decision task consists of employing a collection of $n$ agents, $\mathcal{A}_1, \ldots, \mathcal{A}_n$, that are supposed to cooperate and possibly exchange locally available information. We assume that each agent is described by a triple $\mathcal{A}_i = (\mathcal{S}_i, \mathcal{P}_i, \mathcal{C}_i)$, where $\mathcal{S}_i$ is a collection of sensors, $\mathcal{P}_i$ is a processor that is able to perform elementary logical operations such as $\{\texttt{and}, \texttt{or}, \texttt{not}\}$, and $\mathcal{C}_i$ is a collection of communication devices allowing transmission only of sequences of binary digits, 0 and 1, namely strings of bits.

Although we assume that every agent has the same processing capability, i.e. $\mathcal{P}_i = \mathcal{P}$ for all $i$, we consider situations where agents may be *heterogeneous* in terms of sensors and communication devices. Due to this diversity as well as the fact that agents are placed at different locations, a generic agent $\mathcal{A}_i$ may or may not be able to measure a given input event $u_j$, for $j \in 1, \ldots, m$. We can model this fact by introducing a *visibility matrix* $V \in \mathbb{B}^{n \times m}$ s.t. we have $V(i,j) = 1$ if, and only if, agent $\mathcal{A}_i$ is able to measure the input event $u_j$, or, in other words, if the $i$-th agent is directly *reachable* from the $j$-th input. Moreover, for similar reasons of diversity and for reducing battery consumption, it is reasonable to assume that each agent is able to communicate only with a subset of other agents, which can be captured by introducing a *communication matrix* $C \in \mathbb{B}^{n \times n}$, where $C(i,k) = 1$ if, and only if, agent $\mathcal{A}_i$ is able to receive a data from agent $\mathcal{A}_k$. Agents specified by non-null elements of the row $C(i,:)$ are referred to as $C$-neighbors of the $i$-th agent.

The introduction of visibility relations between inputs and agents immediately implies that, at any instant $t$, only a subset of agents is able to measure the state of each input $u_j$, for all $j$. Therefore, to effectively accomplish the given decision task, we need that such an information "flows" from one agent to another, in a manner that is compliant with the actually available communication paths. On the one hand, this motivates investigation on methods to efficiently aggregate and route information toward a centralized collector. However, we pursue a different yet useful approach where all agents are required to reach a unique network decision on the value of the centralized decision $y = f(u)$, so that any agent can be *polled* and provide consistent complete information. In this perspective, we pose the problem of reaching a *consensus on logical values*.

Furthermore, consider the general fact that any logical function in the centralized decision system $y = f(u)$ can be written as a combination of a minimal set of $q$ subterms, $l_1, \ldots, l_q \in \mathbb{B}$, i.e. each $f_h$ is in the form $f_h = l_1 \oslash_1 l_2 \oslash_2 (\oslash_3(l_3)) \cdots$, where $\oslash_i$ is one of $\{\texttt{and}, \texttt{or}\}$, for $i = 1, 2$, $\oslash_3 = \texttt{not}$, and each subterm may depend only on some of the inputs. Due to its minimality, this formal representation is an *encoding* of the decisions $f$ that is optimal in terms of *specification complexity*, i.e. the number of bits that are necessary to represent all logical functions in $f$. To clarify this, consider the following example with $p = 2$ decisions depending on $m = 2$ input events:

$$\begin{cases} y_1(t) = u_1(t) \bar{u}_2(t), \\ y_2(t) = u_2(t), \end{cases} \quad (3)$$

and assume that $n = 4$ agents have the input visibility described by the following $V$:

$$V = \begin{pmatrix} 1 & 1 \\ 0 & 0 \\ 0 & 1 \\ 1 & 0 \end{pmatrix}. \quad (4)$$

A minimal encoding of the decision system of Eq. 3 is obtained by choosing e.g. $l_1(u) = u_1$, and $l_2(u) = \bar{u}_2$. Indeed, we can write:

$$\begin{cases} y_1(t) = l_1(t) \oslash_1 l_2(t), \\ y_2(t) = \oslash_2(l_2(t)), \end{cases} \quad (5)$$

where we have $\oslash_1 = \texttt{and}$, and $\oslash_2 = \texttt{not}$.

The visibility of each subterm $l_h(u)$ w.r.t. each agent depends on which inputs appear in the subterm itself, and hence another visibility matrix $\hat{V}$ should be introduced to characterize it. Without loss of generality, we will assume that each subterm depends on only one input, so that the same $V$ can be used, i.e. $l_h = \chi_h(u_l)$, for $h = 1, \ldots, q$, where $\chi_h$ are scalar functions, and $l$ is in $\{1, \ldots, m\}$. In the example, we have $l_1(u) = \chi_1(u_1), l_2(u) = \chi_2(u_2)$.

In this view, we can imagine that each agent $\mathcal{A}_i$ stores the values of all subterms into a local *state* vector, $X_i = (X_{i,1}, \ldots, X_{i,q}) \in \mathbb{B}^q$, that is a *string of bits*. In practice we have $X_{i,h} \stackrel{\text{def}}{=} l_h$ for all agents $\mathcal{A}_i$, and all subterms $l_k$. Then, let us denote with $X(t) = (X_1^T(t), \ldots, X_n^T(t))^T \in \mathbb{B}^{n \times q}$ a matrix representing the network state at the discrete time $t$. Each agent $\mathcal{A}_i$ is a dynamic node updating its local state $X_i$ by using a *distributed* logical update function $F_i$ depending on its state, on the state of its $C$-neighbors, and on the reachable inputs, i.e. $X_i(t+1) = F_i(X(t), u(t))$. Moreover, we assume that each agent $\mathcal{A}_i$ is able to produce a logical output decision vector $Y_i = (y_{i,1}, \ldots, y_{i,p}) \in \mathbb{B}^p$ through a suitable distributed logical output function $G$ depending on the local state $X_i$ and on the reachable inputs $u$, i.e. $Y_i(t) = G_i(X_i(t), u(t))$. Let us denote with $Y(t) = (Y_1^T(t), \ldots, Y_p^T(t))^T \in \mathbb{B}^{p \times q}$ a matrix representing the network output at a discrete time $t$. Therefore, the dynamic evolution of the network can be modeled by the following *distributed finite-state iterative system*:

$$\begin{cases} X(t+1) = F(X(t), u(t)), \\ Y(t) = G(X(t), u(t)), \end{cases} \quad (6)$$

where we have $F = (F_1^T, \ldots, F_n^T)^T$, with $F_i : \mathbb{B}^q \times \mathbb{B}^m \to \mathbb{B}^q$, and $G = (G_1^T, \ldots, G_n^T)^T$, with $G_i : \mathbb{B}^q \times \mathbb{B}^m \to \mathbb{B}^p$.

It should be apparent that, given a decision system of the form of Eq. 2, and a minimal encoding, the structure of the output function $G$ is fixed. Indeed, map $G_i$ can readily be obtained by simply replacing those subterms $l_h$ in Eq. 2 that are not visible from agent $\mathcal{A}_i$ with the corresponding state component $X_{i,h}$. To show this, consider again the example of Eq. 3 and Eq. 4. For the given $V$, the outputs of the agents are $Y_i(t) = (y_{i,1}(t), y_{i,2}(t))$ and can be written as follows:

$$\begin{aligned}
Y_1(t) &= G_1(X_1, u) = (u_1(t)\,\bar{u}_2(t), u_2(t))\,, \\
Y_2(t) &= G_2(X_2, u) = (X_{2,1}(t)\,X_{2,2}(t), \bar{X}_{2,2}(t))\,, \\
Y_3(t) &= G_3(X_3, u) = (X_{3,1}(t)\,\bar{u}_2(t), u_2(t))\,, \\
Y_4(t) &= G_4(X_4, u) = (u_1(t)\,X_{4,2}(t), \bar{X}_{4,2}(t))\,.
\end{aligned} \qquad (7)$$

As a matter of fact, the only degree of freedom in the design of a logical consensus is the distributed map $F$ that will be designed based on a given pair $(C, V)$. In this perspective, we are interested in solving the following design problem:

*Problem 1 (Globally Stable Synthesis):* Given a decision system of the form of Eq. 2, a visibility matrix $V$, and a communication matrix $C$, design a logical consensus system of the form of Eq. 6, that is compliant with $C$ and $V$, and s.t., for all initial network state $X(0)$, and all inputs $u$, there exists a finite time $\bar{N}$ s.t. the system reaches a consensus on the centralized decision $y^* = f(u)$, i.e. $Y(t) = \mathbf{1}_n\,(y^*)^T$, for all $t \geq \bar{N}$.

Another important property is the ability of a distributed logical consensus system to tolerate a given number of possible faulty or misbehaving agents. In this perspective we are interested in solving also the following:

*Problem 2 (Robust Design):* Under the same hypotheses of Problem 1, and assuming that at most $\gamma$ agents in a set $\Gamma$ can send *corrupted* data, design a *robust* logical consensus system guaranteeing that all other agents reach an agreement on the correct consensus value, i.e. $Y_i(t) = (y^*)^T$, for all $i \notin \Gamma$, and $t \geq \bar{N}$.

## III. Convergence of Boolean Dynamic Systems

Consider the simplest Boolean algebra described by the sextuple $(\mathbb{B}, +, \cdot, \neg, 0, 1)$, where $\mathbb{B} = \{0, 1\}$ is a domain set, $+$ and $\cdot$ are binary operations representing the logical sum and product, respectively, $\neg$ is a unary operation representing the logical complement, $0$ (null) is the smallest value, and $1$ (unity) is the biggest value of the domain. Consider also a partial order relation $\leq$ between the elements of $\mathbb{B}$ described by the axioms: $0 \leq 0$, $0 \leq 1$, $1 \leq 1$. We recall from [11] and [12] known results on the convergence of an autonomous Boolean dynamic system of the form

$$\begin{cases} x(t+1) = F(x(t))\,, \\ x(0) = x^0\,, \end{cases}$$

where $x = (x_1, \ldots, x_n)^T \in \mathbb{B}^n$ is the system's state, $F : \mathbb{B}^n \to \mathbb{B}^n$ is an application producing a new vector state from the current state's components, by using only the operations $+$, $\cdot$ and $\neg$, and $x^0$ is the system's initial state.

First note that, as $\mathbb{B}^n$ is a finite set, the convergence of the state sequence generated by iterations of $F$ corresponds to the fact that the sequence itself becomes stationary after a certain time $\bar{t}$. Moreover, consider the following definitions:

*Definition 1 (Eigenvalues and Eigenvectors):* A scalar $\lambda \in \mathbb{B}$ is an *eigenvalue* of a Boolean matrix $A \in \mathbb{B}^{n \times n}$ if there exists a vector $x \in \mathbb{B}^n$, called *eigenvector*, s.t.

$$A\,x = \lambda\,x\,.$$

*Definition 2 (Incidence Matrix):* The incidence matrix of a Boolean map $F$ is a Boolean matrix $B(F(x)) = \{b_{i,j}\}$, where $b_{i,j} = 1$ if, and only if, the $i$–th component of $F(x)$ depends on the $j$–th component of the input vector $x$.

*Definition 3 (Boolean Spectral Radius):* The spectral radius of a Boolean matrix $A \in \mathbb{B}^{n \times n}$, denoted with $\rho(A)$, is its biggest eigenvalue in the sense of the vector order relation $\leq$. Consider also the following propositions:

*Proposition 1:* Every Boolean matrix $A \in \mathbb{B}^{n \times n}$ has at least one eigenvalue. Hence $\rho(A)$ always exists.

*Proposition 2:* A Boolean matrix $A \in \mathbb{B}^{n \times n}$ has Boolean spectral radius $\rho(A) = 0$ if, and only if, one of the two following equivalent conditions hold:
- $P^T A P$ is a strictly lower or upper triangular matrix for some permutation matrix $P$;
- $A^n = 0$ (meaning the $n$–th Boolean matrix power of $A$).

We can readily recall from [11] the main result on the global convergence of a Boolean map:

*Theorem 1:* A map $F : \mathbb{B}^n \to \mathbb{B}^n$ globally converges to a unique equilibrium if the following equivalent conditions hold:
- $\rho(B(F(x))) = 0$;
- $P^T B(F(x))\,P$ is strictly lower or upper triangular, for some permutation matrix $P$;
- $B(F(x))^q = 0$, with $0 \leq q \leq n$;
- $\exists q \leq n$ s.t. $F^q$ (the composition of $F$ with itself $q$ times) is a constant map, i.e. it is independent of $x(0)$. ♦

Given an equilibrium point $\bar{x}$ of the Boolean map $F$, i.e. $F(\bar{x}) = \bar{x}$, consider the following definitions:

*Definition 4 (Von–Neumann Neighborhood (VNN)):* Given a point $x \in \mathbb{B}^n$, its VNN is the set $V(x)$ of all points differing from $x$ in at most one component, i.e.

$$V(x) = \{x, \tilde{x}^1, \cdots, \tilde{x}^n\}\,,$$

where $\tilde{x}^j = (x_1, \cdots, x_{j-1}, \neg x_j, x_{j+1}, \cdots, x_n)^T$.
If $\mathbb{B}^n$ is represented as a hypercube and all its elements as its vertices, $\tilde{x}^j$ can be interpreted as the $j$–th vertex adjacent to $x$.

*Definition 5 (Discrete Derivative):* The discrete derivative of a Boolean map $F : \mathbb{B}^n \to \mathbb{B}^n$ at a generic point $x \in \mathbb{B}^n$ is a Boolean matrix $F'(x) = \{F'_{i,j}\}$, s.t. $F'_{i,j} = 1$ if, and only if, a variation in the $j$–th component of $x$ produces a variation in the $i$–th component of $F(x)$, i.e.,

$$F'_{i,j}(x) = F_i(x) \oplus F_i(\tilde{x}^j)\,,$$

where $\oplus$ is the exclusive disjunction

$$\begin{aligned}
\oplus \;:\; & \mathbb{B} \times \mathbb{B} \to \mathbb{B} \\
& (x_i, y_i) \mapsto (\neg x_i\,y_i) + (x_i\,\neg y_i)\,.
\end{aligned}$$

Finally, consider the following two notions:

*Definition 6:* An equilibrium point $x \in \mathbb{B}^n$ is said to be *attractive* in its VNN $V(x)$ if the following two relations hold:
- $F(y) \in V(x)$, for all $y \in V(x)$;

- there exists $n \in \mathbb{N}$ s.t., for all $y \in V(x)$, $F^n(y) = x$.

*Definition 7:* A Boolean map $F$ is said to be *locally convergent* at an equilibrium point $x$ if $x$ is attractive in its VNN. We can recall the main result on the attractiveness of an equilibrium point [12]:

*Theorem 2:* An equilibrium point $x \in \mathbb{B}^n$ is attractive in its VNN if, and only if, the following two relations hold:
- $\rho(F'(x)) = 0$,
- $F'(x)$ contains at most one non–null element in each column. ♦

*Example 3.1:* Consider the map
$$F(x) = \begin{pmatrix} x_3(x_1 + \neg x_2) \\ x_3(x_1 + x_2) + \neg x_3(\neg x_1 + x_2) \\ x_1 \end{pmatrix}.$$

Its incidence matrix is
$$B(F(x)) = \begin{pmatrix} 1 & 1 & 1 \\ 1 & 1 & 1 \\ 1 & 0 & 0 \end{pmatrix},$$

whose spectral radius is $\rho(B(F(x))) = 1$, which tells us that the map is not contractive. Thus, the presence of multiple equilibria or cycles cannot be excluded. Indeed, let us find the equilibria states $\bar{x}$ being s.t. $F(\bar{x}) = \bar{x}$, or equivalently
$$\neg(F_i(\bar{x}) \oplus \bar{x}_i) = 1 \text{ for } i = 1, 2, 3.$$

After some simplification, this gives the Boolean equations $\bar{x}_1\bar{x}_3 + \neg\bar{x}_1\neg\bar{x}_3 = 1$, $\bar{x}_1\bar{x}_3 + \neg\bar{x}_1\neg\bar{x}_2 + \neg\bar{x}_2\bar{x}_3 + \neg\bar{x}_1\neg\bar{x}_3 = 1$, and $\neg\bar{x}_1(\bar{x}_2 + \bar{x}_3) + \bar{x}_2 + \bar{x}_3 = 1$, which are solved by the vectors
$$\bar{x}^{(1)} = (0, 1, 0)$$
$$\bar{x}^{(2)} = (1, 1, 1).$$

Moreover, the discrete derivatives of the map $f$ at the two equilibria are
$$F'(\bar{x}^{(1)}) = \begin{pmatrix} 0 & 0 & 0 \\ 0 & 0 & 0 \\ 1 & 0 & 0 \end{pmatrix}, \quad F'(\bar{x}^{(2)}) = \begin{pmatrix} 1 & 0 & 1 \\ 0 & 0 & 0 \\ 1 & 0 & 0 \end{pmatrix},$$

which tell us, based on Theorem 2, that the first equilibrium in $\bar{x}^{(1)}$ is attractive in its VNN, whereas the second one in $\bar{x}^{(2)}$ is not. ♦

Finally, we can provide the following definition:

*Definition 8:* A logical map $F : \mathbb{B}^n \times \mathbb{B}^m \to \mathbb{B}^n$ is $(C, V)$–*compliant* if, and only if, its incidence matrix $B(F(X, u))$ w.r.t. the state and the inputs satisfies the logical vector inequality
$$B(F(X, u)) \leq (C|V).$$

## IV. DISTRIBUTED MAP SYNTHESIS FOR LOGICAL CONSENSUS

In this section a solution for Problem 1 is presented consisting of an algorithm that generates an optimal distributed logical linear consensus system. More precisely, the algorithm produces a $(C, V)$–compliant linear iteration map $F$ minimizing the number of messages to be exchanged, and the time needed to reach a consensus (a.k.a. *rounds*).

To achieve this we first need to understand how the agent network can reach a consensus on the value of the $j$–th subterm $l_j$ in the decision system of Eq. 2. Without loss of generality, let us pose $l_j = u_j$ and consider the $j$–th column $V_j$ of the visibility matrix $V$ that also describes the visibility of $l_j$. We need a procedure for finding to which agents the value of input $u_j$ can be propagated. First note that vector $V_j$ contains 1 in all entries corresponding to agents that are able to "see" $u_j$, or, in other words, it specifies which agents are directly *reachable* from $u_j$. Then, it is useful to consider vectors $C^k V_j$, for $k = 0, 1, \ldots$, each containing 1 in all entries corresponding to agents that are reachable from input $u_j$ after *exactly* $k$ steps. The $i$–th element of $C^k V_j$ is 1 if, and only if, there exists a *path* of length $k$ from any agent directly reached by $u_j$ to agent $\mathcal{A}_i$. Recall that, by definition of graph diameter, all agents that are reachable from an initial set of agents are indeed reached in at most $\text{diam}(G)$ steps, with $\text{diam}(G) \leq n-1$. Let us denote with $\kappa$ the *visibility diameter* of the pair $(C, V_j)$ being the number of steps after which the sequence $\{C^k V_j\}$ does not reach new agents. Thus, given a pair $(C, V_j)$, we can conveniently introduce the following *reachability matrix* $R_j$, assigned with input $u_j$:

$$R_j = \begin{pmatrix} V_j & CV_j & C^2V_j & \cdots & C^{n-1}V_j \end{pmatrix}, \quad (8)$$

whose columns *span* a subgraph $G_\mathcal{R}(N_\mathcal{R}, E_\mathcal{R})$ of $G$, where $N_\mathcal{R}$ is a node set of all agents that are *eventually* reachable from input $u_j$, and $E_\mathcal{R}$ is an unspecified edge set, that will be considered during the design phase. Computing the span of $R_j$ is very simple and efficient, and indeed all reachable agents, that are nodes of $N_\mathcal{R}$, are specified by non–null elements of the Boolean vector $I_j = \sum_{k=0}^{n-1} C^k V_j = \sum_{k=0}^{n-1} R_j(:, i)$, that is the logical sum of all columns in $R_j$ and that contains 1 for all agents for which there exists at least one path originating from an agent that is able to measure $u_j$. Then, we can partition the agent network into $N_\mathcal{R} = \{i \mid I_j(i) = 1\}$, and $N_{\bar{\mathcal{R}}} = N \setminus N_\mathcal{R}$, where $N = \{1, \ldots, n\}$. In this perspective we can give the following:

*Definition 9:* A pair $(C, V_j)$ is *(completely) reachable* if, and only if, the corresponding reachability matrix $R_j(C, V_j)$ spans the entire graph, i.e. $N_\mathcal{R} = N$.

Consider e.g. a network with $n = 5$ agents, and the following pair of communication and visibility matrices:

$$C = \begin{pmatrix} 1 & 1 & 0 & 0 & 1 \\ 1 & 0 & 1 & 0 & 1 \\ 1 & 1 & 1 & 1 & 1 \\ 0 & 1 & 1 & 1 & 1 \\ 0 & 0 & 0 & 0 & 1 \end{pmatrix} \quad V_j = \begin{pmatrix} 1 \\ 0 \\ 0 \\ 0 \\ 0 \end{pmatrix}. \quad (9)$$

Observe that only agent $\mathcal{A}_1$ is able to measure $u_j$. The $j$–th reachability matrix $R_j = (V_j \ CV_j \ C^2V_j \ C^3V_j \ C^4V_j)$ is:

$$R_j = \begin{pmatrix} 1 & 1 & 1 & 1 & 1 \\ 0 & 1 & 1 & 1 & 1 \\ 0 & 1 & 1 & 1 & 1 \\ 0 & 0 & 1 & 1 & 1 \\ 0 & 0 & 0 & 0 & 0 \end{pmatrix}. \quad (10)$$

Simple computation gives $I_j = (1, 1, 1, 1, 0)^T$ which readily reveals that $N_\mathcal{R} = \{1, 2, 3, 4\}$ is the node set of the reachable subgraph, and $N_{\bar{\mathcal{R}}} = \{5\}$ is the node set of the unreachable subgraph. Furthermore, all agents in $N_\mathcal{R}$ are reached by input $u_j$ within $\kappa = 3$ steps.

The design phase can obviously concern only the reachable subgraph $G_\mathcal{R}(N_\mathcal{R}, E_\mathcal{R})$ of $G$, and in particular will determine the edge set $E_\mathcal{R}$. Moreover, observe that a non–empty unreachable subgraph $G_{\bar{\mathcal{R}}}$ in a consensus context is a symptom of the fact that the design problem is not well–posed. In practice, this would require changing sensor's visibility and locations in order to have a reachable $(C, V_j)$ pair.

Let us suppose, as in the example, that only agent $\mathcal{A}_1$ is able to measure $u_j$. Then, a straightforward and yet optimal *strategy to allow the information on $u_j$ flowing* through the network is obtained if agent $\mathcal{A}_1$ communicates its measurement to all its $C$–neighbors, which in turn will communicate it to all their $C$–neighbors without overlapping, and so on. In this way, we have that every agent $\mathcal{A}_i$ receives $u_j$ from exactly one minimum–length path originating from agent $\mathcal{A}_1$. The vector sequence $\{C^k V_j\}$ can be exploited to this aim. Indeed, it trivially holds that $C^k V_j = C(C^{k-1} V_j)$, meaning that agents reached after $k$ steps have received the input value from agents that were reached after exactly $k-1$ steps. Then, any consecutive sequence of agents that is extracted from non–null elements of vectors in $\{C^k V_j\}$ are $(C, V_j)$–compliant by construction. A consensus strategy would minimize the number of rounds if, and only if, at the $k$-th step, all agents specified by non–null elements of vector $C^k V_j$ receives the value of $u_j$ from the agents specified by non–null elements of vector $C^{k-1} V_j$. Nevertheless, to minimize also the number of messages, only agents specified by non–null elements of vector $C^k V_j$ and that have not been reached yet must receive $u_j$. If vector $I_j = \sum_{i=0}^{i=k} C^i V_j$ is iteratively updated during the design phase, then the set of all agents that must receive a message on $u_j$ are specified by non–null elements of vector $C^k V_j \wedge \neg I_j$. By doing this, an optimal pair $(C^*, V_j^*)$ allowing a consensus to be established over the reachable subgraph is obtained. In the considered example, we have:

$$C^* = \begin{pmatrix} 0 & 0 & 0 & 0 & 0 \\ \hline 1 & 0 & 0 & 0 & 0 \\ 1 & 0 & 0 & 0 & 0 \\ \hline 0 & 1 & 0 & 0 & 0 \\ \hline 0 & 0 & 0 & 0 & 0 \end{pmatrix}, \quad V^* = \begin{pmatrix} 1 \\ \hline 0 \\ 0 \\ \hline 0 \\ \hline 0 \end{pmatrix}. \quad (11)$$

Observe that is $C^* = S\, C \leq C$, where $S$ is a suitable selection matrix.

This procedure actually gives us only a suggestion on the incidence matrix of the iteration map $F(X, u)$ w.r.t. the state $X$ and the input $u$. More precisely, it requires that $\mathcal{B}(F(X, u)) = (C^* | V_j^*)$. Nevertheless, we prove in following Theorem 3 that a simple logical linear consensus algorithm of the form

$$x(t+1) = F_j\, x(t) + B_j\, u_j(t), \quad (12)$$

where $F_j = C^*$, $B_j = V_j^*$, and $x \in \mathbb{B}^n$, allows a consensus to be reached through the entire reachable subgraph. The state $x$ must be interpreted as the network *distributed estimation* of the value of the subterm $l_j$ or $u_j$. It is indeed a vector and not a matrix, since we are concerned here only with the $j$-th input. In our example, we have:

$$\begin{cases} x_1(t+1) = u(t), \\ x_2(t+1) = x_1(t), \\ x_3(t+1) = x_1(t), \\ x_4(t+1) = x_2(t). \end{cases} \quad (13)$$

In all cases where a unique generic agent $\mathcal{A}_i$ is directly reachable from input $u_j$, an optimal communication matrix $C^*$ for a linear consensus of the form of Eq. 12 can be iteratively found as the incidence matrix of a *input–propagating spanning tree* having $\mathcal{A}_i$ as the root. Then, an optimal pair $(C^*, V_j^*)$ can be written as $C^* = P^T (S\, C)\, P$, and $V_j^* = P^T V_j$, where $S$ is a selection matrix, and $P$ is a permutation matrix. Furthermore, $C^*$ has the following lower–block triangular form:

$$C^* = \begin{pmatrix} 0 & 0 & \cdots & 0 & 0 \\ \tilde{C}_{i,1} & 0 & \cdots & 0 & 0 \\ \vdots & & & \vdots & \vdots \\ 0 & \cdots & \tilde{C}_{i,\kappa_i} & 0 & 0 \\ \hline 0 & \cdots & 0 & 0 & 0 \end{pmatrix}, \quad (14)$$

and $V_j^* = P^T V_j = (1, 0, \ldots, 0)^T$.

The optimal pair $(C^*, V_j^*)$ preserves the reachability property of the original pair $(C, V_j)$. Direct computation of the reachability matrix $R_j^*$ yields:

$$R_j^* = \begin{pmatrix} 1 & 0 & 0 & \cdots & 0 & 0 & 0 \\ 0 & \tilde{C}_{i,1} & 0 & \cdots & 0 & 0 & 0 \\ 0 & 0 & \tilde{C}_{i,2}\tilde{C}_{i,1} & \cdots & 0 & 0 & 0 \\ \vdots & & & & & \vdots & \vdots \\ 0 & 0 & 0 & \cdots & 0 & \tilde{C}_{i,\kappa_i}\cdots\tilde{C}_{i,1} & 0 \\ \hline 0 & 0 & 0 & \cdots & 0 & 0 & 0 \end{pmatrix},$$

where the upper–left matrix block, that is related to the reachable subgraph, contains exactly a 1 in all rows, whereas the two lower matrix blocks are 0, and indeed they are related to the unreachable subgraph. Finally, observe that all products $\tilde{C}_{i,j}\tilde{C}_{i,j+1}$ are well–defined since the column number of $\tilde{C}_{i,j}$ equals the row number of $\tilde{C}_{i,j+1}$ by construction.

We are now ready to consider the more general case with $\nu$, $1 \leq \nu \leq n$ agents that are reachable from input $u_j$, and let us denote with $A = \{i_1, \ldots, i_\nu\}$ the index set of such agents. Then, the optimal strategy for propagating input $u_j$ consists of having each of the other agents receive the input measurement through a path originating from the nearest reachable agent in $A$. This naturally induces a network partition into $\nu$ disjoint subgraphs or spanning trees, each directly reached by the input through a different agent. Let us extract $\nu$ independent vectors $V_j(i_1), \ldots, V_j(i_\nu)$ from vector $V_j$ having a 1 in position $i_h$. Then, the sequences $\{C^k V_j(i_h)\}$ are to be considered to compute the optimal partition. Let us denote with $\kappa_i$, for all $i \in A$ the number $k$ of steps for the sequence $\{C^k V_j(i)\}$ to become stationary. Therefore, we have that the visibility diameter of the pair $(C, V_j)$ is vis-diam$(C, V_j) = \max_i \{\kappa_i\}$. Without loss of generality, we can image that $\kappa_1 \geq \kappa_2 \geq \cdots \geq \kappa_\nu$. Therefore, for the generic case, there exist a permutation matrix $P$ and a selection matrix $S$ s.t. an optimal pair $(C^*, V_j^*)$ can be obtained as

$C^* = P^T (S C) P$, $V_j^* = P^T V_j$, where

$$C^* = \text{diag}(C_1, \ldots, C_\nu), V_j^* = (V_{j,1}^T, \ldots, V_{j,\nu}^T)^T, \quad (15)$$

and where each $C_i$ and $V_{j,i}$ have the form of the Eq. 14. Finally, the actual optimal linear consensus algorithm is obtained choosing $F_j = P C^*$, and $B_j = P V_j^*$. Consider e.g. a network of $n = 5$ agents and the following pair $(C, V_j)$ with $\nu = 2$:

$$C = \begin{pmatrix} 1 & 1 & 0 & 0 & 1 \\ 1 & 0 & 1 & 0 & 1 \\ 0 & 1 & 1 & 1 & 1 \\ 0 & 1 & 1 & 1 & 1 \\ 1 & 0 & 1 & 0 & 1 \end{pmatrix} V_j = \begin{pmatrix} 1 \\ 1 \\ 0 \\ 0 \\ 0 \end{pmatrix}. \quad (16)$$

An optimal pair $(C^*, V_j^*)$ allowing a consensus to be established over the complete graph $G$ is given by

$$C^* = \left(\begin{array}{c|cc|cc} 0 & 0 & 0 & 0 & 0 \\ \hline 1 & 0 & 0 & 0 & 0 \\ 1 & 0 & 0 & 0 & 0 \\ \hline 0 & 0 & 0 & 0 & 0 \\ 0 & 0 & 0 & 1 & 0 \end{array}\right), V_j^* = \left(\begin{array}{c} 1 \\ \hline 0 \\ 0 \\ \hline 1 \\ 0 \end{array}\right).$$

The corresponding optimal linear consensus algorithm is:

$$\begin{cases} x_1(t+1) = u(t), \\ x_2(t+1) = u(t), \\ x_3(t+1) = x_2(t), \\ x_4(t+1) = x_2(t), \\ x_5(t+1) = x_1(t). \end{cases} \quad (17)$$

Algorithm 1 allows computation of an optimal pair $(C^*, V_j^*)$ as in Eq. 15. Its asymptotic *computational complexity* is in the very worst case $O(n^2)$, where $n$ is the number of agents, and its *space complexity* in terms of memory required for its execution is $\Omega(n)$. However, its implementation can be very efficient since it is based on Boolean operations on bit strings. Finally, *communication complexity* of a run of the consensus protocol in terms of the number of rounds is $\Theta(\texttt{vis-diam}(C, V_j))$.

To conclude, we need to prove that a so–built logical consensus system does indeed solve Problem 1. Hence, for the general case with $\nu \geq 1$ agents that are reachable from input $u_j$, we can the state the following:

*Theorem 3 (Global Stability of Linear Consensus):* A logical linear consensus system of the form $x(t+1) = C^* x(t) + V_j^* u_j(t)$, where $C^*$ and $V_j^*$ are obtained as in Eq. 15 from a reachable pair $(C, V_j)$, converges to a unique network agreement given by $\mathbf{1}_n u_j$ in at most $\texttt{vis-diam}(C, V_j)$ rounds.

*Proof:* The unique equilibrium of the consensus system is $\mathbf{1}_n u_j$. Indeed, we can focus on the $i$–th subsystem of all agents that directly or indirectly receive $u_j$ from agent $\mathcal{A}_i$. Let us denote with $x_i$ the subsystem state and with $x_{i,j}$ its

---

**Algorithm 1** Optimal Linear Synthesis by Input–Propagation

**Inputs:** $C$, $V_j$
**Outputs:** Minimal pair $(C^*, V_j^*)$, permutation $P$.

Set $k \leftarrow 0$
Set $I = \mathbf{diag}(V_j)$
Set $N = \neg V_j$ ◁ nodes not yet reached
**repeat**
  Set $k \leftarrow k + 1$
  **for all** nodes $i$ **do**
    Set $Adj_{(:,i)} = C_{(:,i)}^k \wedge N$ ◁ boolean vector of new nodes reached at the $k$–steps from node $i$
    Set $I_{(:,i)} = I_{(:,i)} \vee Adj_{(:,i)}$ ◁ boolean vectors of nodes reached in at most $k$–step from node $i$
    Set $N = N \wedge \neg Adj_{(:,i)}$ ◁ boolean vector of nodes not yet reached
    Set $\tilde{C}_{(:,i)} = C_{(:,i)} \wedge Adj_{(:,i)}$
  **end for**
**until** $(Adj \neq 0^{n \times n}) \wedge (N \neq 0^{n \times 1})$
Compute $\kappa_i = \text{card}(I_{(:,i)})$, $i = 1, \ldots, n$
Compute $I = \text{sort}(I_{(:,i)} \mid \kappa_i > \kappa_{i+1})$
Set $P = \sum_i \mathbf{diag}(I_{(:,i)})$
Set $C^* = P^T \tilde{C} P$
Set $V_j^* = P^T V_j$

---

components. Then, the subsystem dynamics is:

$$\begin{cases} x_{i,0}(t+1) &= u(t), \\ x_{i,1}(t+1) &= \tilde{C}_{i,1} \, x_{i,0}(t), \\ &\vdots \\ x_{i,l}(t+1) &= \tilde{C}_{i,l} \, x_{i,l-1}(t), \\ &\vdots \\ x_{i,\kappa_i}(t+1) &= \tilde{C}_{i,\kappa_i} \, x_{i,\kappa_i-1}(t). \end{cases} \quad (18)$$

The system of equations has a strictly lower–triangular form, and hence it can iteratively be solved block–wise as in the Gauss' method. Indeed, the first row gives the scalar relation $x_{i,0}(t) = u(t)$. Then, the second row is $x_{i,1}(t) = \tilde{C}_{i,1} x_{i,0}(t)$, where $\tilde{C}_{i,1}$ is a vector with all entries equal to 1, that correspond to a set of agents that are updated after 1 step. In particular we have: $x_{i,1}(t) = x_{i,0}(t) = u$. At the generic iteration $k$, a block of variables $x_{i,k}$ are updated through the $k$–th matrix $\tilde{C}_{i,k}$ having exactly a 1 in each row. Hence we have $x_{i,k}(t) = x_{i,k-1}(t) = u$. After at most $\kappa_i$ steps, all agents in the $i$-th subgraph are set to $x_i^* = \mathbf{1}_{n_i} u$, where $n_i$ is the number of agents of the considered subgraph itself. By repeating this procedure for all $\nu$ blocks, and since all agents that are directly reachable from input $u_j$ read the same value $u_j$, we can prove that the entire network reaches an agreement on the unique global equilibrium $x^* = \mathbf{1}_n u$ in at most $\max\{\kappa_1, \ldots, \kappa_\nu\} = \texttt{vis-diam}(C, V_j)$ steps. ∎

## V. SENSOR FAILURE AND ROBUST MAP DESIGN

In this section a solution for Problem 2 is provided by dealing with possible sensor failure that may lead the consensus system of Eq. 6 to reach incorrect global decisions, such as raising false alarms in an intrusion detection application. We begin by considering a *temporary fault* that occurs whenever measurements of one or more agents are corrupted by noise. This type of fault can be modeled as inconsistent variations of the initial state $x(0)$. Hence, when dealing with temporary faults, we have to consider the stability of the equilibrium $1_n u_j$ and its basin of attraction. In the previous section we have shown that consensus systems synthesized according to Algorithm 1 are globally stable, i.e. the basin of attraction of the equilibrium point $1 u_j$ is the entire reachable subgraph, and thus temporary *false alarms* are canceled out by the system itself.

The problem becomes more difficult in case of *permanent faults* that occur whenever one or more agents are damaged, due to e.g. a spontaneous failure, or even tampering, and do not correctly execute the consensus algorithm. Consider e.g. the consensus algorithm of Eq. 17 and assume that agent $\mathcal{A}_2$ outputs 1 instead of 0 when the input is $u_j = 0$. This will break the network into two *disagreeing parts*. Within such a scenario, we can require that every agent except $\mathcal{A}_2$ will continue to work properly and eventually reach the correct consensus. This motivates the development of techniques for synthesizing *robust* logical consensus systems. Suppose that a maximum number of $\gamma \in \mathbb{N}$ faulty agents have to be tolerated. The key to solve such a problem is in *redundancy* of input measurement and communication. Intuitively, a minimum number $r$ of such sensors must be able to measure the $j$-th input $u_j$ and/or confirm any transmitted data $x$ on $u_j$. Indeed, we are concerned with the following:

*Definition 10 (Rechability with redundancy):* An agent $\mathcal{A}_i$ is said to be reachable from input $u_j$ *with redundancy* $r \in \mathbb{N}$, or shortly $r$–reachable, if, and only if, it can receive at least $r$ measurements of $u_j$ between a direct measurement of $u_j$ and messages of other agents that are $r$–reachable from $u_j$.

Then, *redundant minimum–length paths* are to be found s.t. information on $u_j$ can robustly flow through the network. Again, such paths can be found by considering successive powers of $C^k V_j(i_h)$. Recall that agents directly reachable from $u_j$ are specified by non–null elements of vector $V_j$. Let us denote with $A = \{i_1, i_2 \ldots\}$ the set of these agents. Moreover, agents that are $r$–reachable after 1 step are those that can receive a message from at least $c \geq r$ agents in $A$. In this perspective, we can say that the information flow is "secured" by $c$ supporting messages. These agents are specified by non–null elements of all possible vectors of the form $C V_j(i_1) \cdots C V_j(i_h)$, with $i_j \in A$. According to the same reasoning, agents that are $r$–reachable after $k$ steps are specified by non–null elements of vectors of the form $C^{\alpha_{i_1}} V_j(i_1) \cdots C^{\alpha_{i_r}} V_j(i_r)$, for some $\alpha_{i_j} = 0, 1 \ldots, k$. It is possible to avoid multiple calculation of such vectors, although we will omit this for the sake of space. By iterating the procedure, an optimal pair $(C^*, V_j^*)$ can be obtained s.t. $C^* = P^T (S C) P$, and $V_j^* = P^T V$, where $P$ is a permutation matrix, and $S$ is a selection matrix. Furthermore, the optimal pair $C^*$ has a strictly lower triangular form and $V^*$ has 1 in the first $r$ rows. Indeed we have:

$$C^* = \begin{pmatrix} 0 & 0 & \cdots & 0 & 0 \\ \tilde{C}_{1,1} & 0 & \cdots & 0 & 0 \\ \tilde{C}_{2,1} & \tilde{C}_{2,2} & \cdots & 0 & 0 \\ \vdots & & & \vdots & \vdots \\ \tilde{C}_{\kappa-1,1} & \cdots & \tilde{C}_{\kappa-1,\kappa-1} & 0 & 0 \\ \hline 0 & \cdots & 0 & 0 & 0 \end{pmatrix}, \quad (19)$$

and $V = (1_r^T 0 0 \cdots 0 \,|\, 0)^T$, where $\kappa$ has the role of a *joint visibility diameter*. Each rows of $C^*$ contains exactly $r$ entries equal to 1. Then, consider the $r$–*reachability matrix* that is defined as $R_j^{(r)} = (V_j^*, C^* V_j^*, \cdots, C^{*n-1} V_j^*)$, whose columns span the $r$–reachable subgraph. If $R_j^{(r)}$ spans the entire graph, then the pair $(C, V_j)$ is said to be (completely) $r$–*reachable*.

Provided that $(C, V_j)$ is $r$–reachable, we need to find a suitable distributed *nonlinear* iteration map $F$ allowing each agent to combine its measurement with the one of its $C$–neighbors. The constraint on global stability has to be relaxed, albeit the basin of attraction of the equilibrium $1_n u_j$ has to be large enough to tolerate $\gamma$ faults. First a generic agent $\mathcal{A}_i$ that can measure $u_j$ can use the local rule $F_i = u_j$ to update its state $x_i$, whereas, for all other agents, an optimal choice is obtained as the logical sum of $\gamma + 1$ terms consisting of the logical product of $\gamma + 1$ received state $x_{i_j}$, i.e.

$$F_i = \sum_{l=1}^{\gamma+1} s_l, \; s_l = x_{i_1} \cdots x_{i_{\gamma+1}}. \quad (20)$$

Refer to Algorithm 2 for a procedure to compute such a (nonlinear) robust consensus system. Furthermore, to be able to construct such terms, we can prove that it is necessary and sufficient to have

$$r = 2\gamma + 1, \quad (21)$$

but we omit this proof for the sake of space.

Consider e.g. a scenario with $n = 5$ agents, $\gamma = 1$ possible faults. The required redundancy is $r = 3$, and hence consider following the communication $C$ and visibility $V$ matrices:

$$C = \begin{pmatrix} 1 & 0 & 1 & 0 & 0 \\ 0 & 1 & 1 & 1 & 0 \\ 1 & 1 & 1 & 1 & 0 \\ 0 & 0 & 0 & 0 & 0 \\ 1 & 1 & 1 & 0 & 1 \end{pmatrix}, \; V_j = \begin{pmatrix} 1 \\ 1 \\ 0 \\ 1 \\ 0 \end{pmatrix}.$$

The network is completely reachable. Then, an optimal distributed nonlinear iteration rule $F$ can be obtained by executing Algorithm 2 as follows:

$$\begin{cases} x_1(t+1) = u(t), \\ x_2(t+1) = u(t), \\ x_3(t+1) = x_1(t) x_2(t) + x_1(t) x_4(t) + x_2(t) x_4(t), \\ x_4(t+1) = u(t), \\ x_5(t+1) = x_1(t) x_2(t) + x_1(t) x_3(t) + x_2(t) x_3(t). \end{cases}$$

Finally, we discuss the robustness properties of such a nonlinear logical map $F$ in the following:

*Theorem 4 (Optimal Robust Nonlinear Consensus):* Consider an $r$–reachable pair $(C, V_j)$ satisfying Eq. 21. A

**Algorithm 2** Optimal Robust Synthesis

**Inputs:** $C, V_j, \gamma$
**Outputs:** Optimal $F(x, u_j)$
1: Set $A \leftarrow \{i \,|\, V_j(i) = 1\}$, $I \leftarrow 1$ for all $i \in A$
2: Set $N \leftarrow \{1, \ldots, n\} \setminus I$, $\kappa \leftarrow 0$
3: **for all** $i \in A$ **do** Set $F_i(x, u_j) \leftarrow u_j$ **end for**
4: **repeat**
5:     Find $\{i_1, \ldots, i_r\} \leftarrow \{i \,|\, I(i) = 1\}$    ◁ $r$–reachable agents
6:     $Adj \leftarrow Ce_{i_1} \cdots Ce_{i_r} \wedge \neg I \wedge N$    ◁ new nodes
7:     Find $K \leftarrow Comb(I, Adj, \gamma)$    ◁ combinations of $\gamma + 1$ agents
    (at least 1 from $Adj$ and others from $I$)
8:     Set $I \leftarrow I \vee Adj$, $N \leftarrow N \wedge \neg Adj$
9:     Compute $\mathcal{I} \leftarrow \{h : Adj = 1\}$    ◁ index list
10:     **for all** new nodes $h \in \mathcal{I}$ **do**
11:         **for all** $s_l = x_{i_1} \cdots x_{i_r}$ with $i_j \in K$ **do**
12:             Set $F_h \leftarrow F_h + s_l$
13:         **end for**
14:     **end for**
15:     Set $\kappa \leftarrow \kappa + 1$
16: **until** $N \neq \emptyset$

distributed logical nonlinear consensus system of the form $x(t+1) = F(x(t), u_j(t))$, where rows of $F$ are computed as in Eq. 20, has the unique equilibrium $\mathbf{1}_n u_j$, and its basin of attraction is able to tolerate up to $\gamma$ permanent faults.

*Proof:* The proof of the existence and uniqueness of the equilibrium point $x^* = \mathbf{1}_n u_j$ follows on the same line of Theorem 3 and can be omitted. Furthermore, direct computation of the discrete derivative of the iteration map $F$ at all points $x$ in the Von Neumann neighborhood of the equilibrium point $x^*$ yields: $\frac{\partial}{\partial x} F(x, u_j) = 0_{n \times n}$, that trivially satisfies Theorem 2. Indeed, this result holds for all points that differ in at most $\gamma$ components from $x^*$, thus proving the attractiveness of $x^*$ and the thesis. ∎

## VI. EXAMPLE

Consider an indoor environment $\mathcal{W}$, with a number $n$ of agents or observers $\mathcal{A}_i$ whose task is to detect and locate possible intruders in $\mathcal{W}$. We assume that agents have sensors with star–shaped visibility regions that define a partition of the environment $\mathcal{W}_i$, $i = 1, \ldots, m$ (hence, $\cup_{i=1}^{n} \mathcal{W}_i = \mathcal{W}$ and $\mathcal{W}_i \cap \mathcal{W}_j = \emptyset$, $i \neq j$). The presence or the absence of an intruder in region $\mathcal{W}_j$ can be seen as an input $u_j$ to the following system of $p = m$ logical decisions: $y_i(t) = u_i(t)$, $i = 1, \ldots, m$, that each agent is required to estimate. However, agents are able to detect the presence of intruders only within their visibility areas, which is described by a visibility matrix $V \in \mathbb{B}^{n \times m}$, with $V_{i,j} = 1$ if, and only if, an intruder in region $\mathcal{W}_j$ can be seen by agent $\mathcal{A}_i$. Moreover, let $X \in \mathbb{B}^{n \times m}$ denote the alarm state of the system: $X_{i,j} = 1$ if agent $\mathcal{A}_i$ reports an alarm about the presence of an intruder in region $\mathcal{W}_j$. The alarm can be set because an intruder is actually detected by the agent itself, or because of communications with neighboring observers. Indeed, agents have communication devices that allows them to share alarm states with all other agents that are within *line–of–sight* or nearby. In this context, our objective is to design a consensus mechanism as a distributed update rule $F : \mathbb{B}^n \times \mathbb{B}^m \to \mathbb{B}^n$ of the form $X(t+1) = F(X(t), u(t))$, with $u = (u_1, \ldots, u_m)^T$, whose goal is to lead all observers or agents to achieve the same state ($X_{i,j} = X_{k,j} \; \forall i, k$ and $\forall j$), so that any observer can provide consistent complete information on the environment when polled. In other terms, *at consensus*, each column of $X$ should have either all zeros or all ones. This implies that the consensus update law should have only fixed points in $0 \in \mathbb{B}^n$ or $1 \in \mathbb{B}^n$, depending on the corresponding column of $\mathbf{1}_n f(u) = \mathbf{1}_n u$.

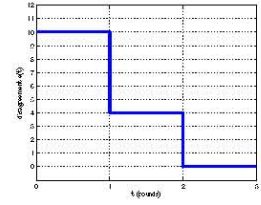

Figure 2. The total network disagreement $e(t)$ relative to the centralized decision $f(u)$ converges to zero in `vis-diam`$(C, V) = 2$ steps.

Consider first applying Algorithm 1 that produces a linear logical consensus of the form $X(t+1) = F X(t) + B u(t)$, where each row basically expresses the rule that an observer alarm is set at time $t+1$ if it sees an intruder (through $u$), or if one of its $C$–neighbors was set at time $t$. The visibility diameter of this pair $(C, V)$ is 2, which will correspond to the maximum number of steps before consensus is reached. Fig. 1 shows snapshots from a typical run of this linear consensus algorithm where every agents converge to consensus after 2 steps, whereas Fig. 2 is a plot of the total *network disagreement* $e(t)$ w.r.t. the centralized decision $f(u)$, i.e. $e(t) = \sum_{j=1}^{m} \sum_{i=1}^{n} X_{i,j}(t) \oplus f_i(u)$, where $\oplus$ is the `xor` operator. If all agents correctly set their alarm states, the system correctly converges to a state where all columns of $X$ are either zero or one. However, this system is not robust to permanent faults (see Fig. 3). A more conservative mechanism can be obtained by applying Algorithm 2, with $\gamma = 1$, that generates a nonlinear rule requiring that agent $\mathcal{A}_i$ sets an alarm regarding $\mathcal{W}_j$ at time $t+1$ if at least two neighboring sensors having visibility on $\mathcal{W}_j$ are in alarm at time $t$, or if it sees an intruder (through $u$). By means of this second system, false alarms raised by at most $\gamma = 1$ misbehaving agents are correctly handled.

## VII. CONCLUSION

In this work we introduced a novel consensus mechanism where agents of a network are able to share logical values and proposed an algorithm that produces optimal logical linear consensus systems. To cope with possible sensor failures, we also presented an algorithm by which robust logical consensus systems can be built. We applied this to a case study consisting of a distributed IDS. While the execution of the linear or non-linear consensus systems presented above are distributed, their design requires complete knowledge of the communication graph, a limitation that will be addressed in future work.

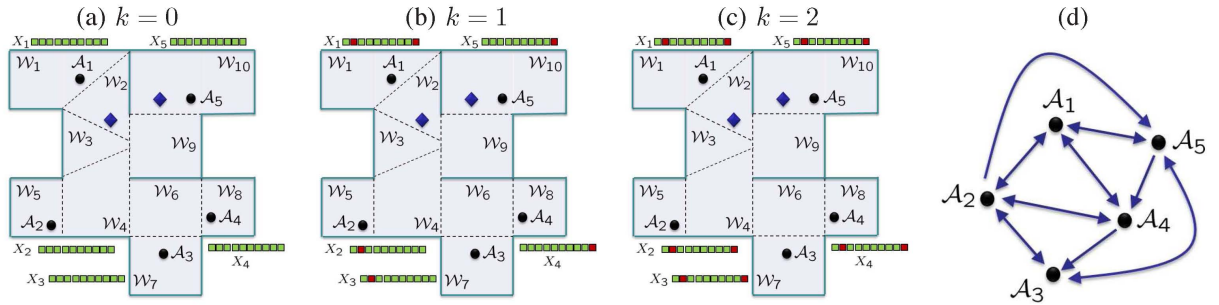

Figure 1. (a)–(c) Run of the linear consensus system with 2 intruders (blue rhombus) in regions $\mathcal{W}_2$ and $\mathcal{W}_{10}$, respectively. The figure sequence shows that a correct agreement is reached (components of the state $X_i$ of every agents are green or 0, when no intruder is detected in the corresponding region, red or 1 otherwise). (d) Considered communication graph $C$.

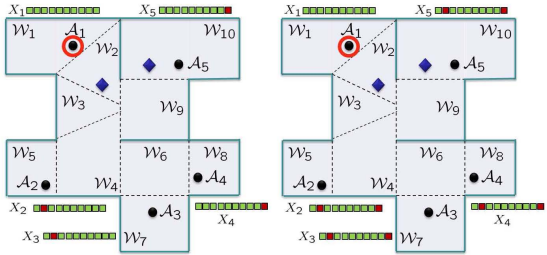

Figure 3. Final network decisions in case of permanent fault of $\mathcal{A}_1$ that incorrectly sets its state to 0. An agreement is not reached by means of the linear consensus system (left), whereas this misbehavior is tolerated by the nonlinear ones (right).

ACKNOWLEDGMENT

This work has been partially supported by the European Commission with contract FP7-IST-2008-224428 "CHAT - Control of Heterogeneous Automation Systems: Technologies for scalability, reconfigurability and security", with contract number FP7-2007-2-224053 CONET, the "Cooperating Objects Network of Excellence", with contract number FP7-2010-257649 PLANET, "PLAtform for the deployment and operation of heterogeneous NETworked cooperating objects", and with contract number FP7-2010-257462 HYCON2, "Highly-complex and networked control systems".

REFERENCES

[1] R. Olfati-Saber, J. A. Fax, and R. N. Murray, "Consensus and Cooperation in Networked Multi–Agent Systems," *Proc. of the IEEE*, 2007.

[2] W. Ren, R. Beard, and E. Atkins, "Information consensus in multivehicle cooperative control," *IEEE Cont. Syst. Mag.*, vol. 27, no. 2, pp. 71–82, 2007.

[3] V. Blondel, J. Hendrickx, A. Olshevsky, and J. Tsitsiklis, "Convergence in Multiagent Coordination, Consensus, and Flocking," *IEEE Int. Conf. on Decision and Control*, pp. 2996–3000, 2005.

[4] L. Fang, P. Antsaklis, and A. Tzimas, "Asynchronous Consensus Protocols: Preliminary Results, Simulations and Open Questions," *IEEE Int. Conf. on Decision and Control and Eur. Control Conference*, pp. 2194–2199, 2005.

[5] D. Bertsekas and J. Tsitsiklis, "Parallel and Distributed Computation: Numerical Methods," 2003.

[6] N. Lynch, *Distributed Algorithms*. Morgan Kaufmann Publishers.

[7] J. Cortés, "Distributed algorithms for reaching consensus on general functions," *Automatica*, 2007.

[8] K. Marzullo, "Maintaining the time in a distributed system: An example of a loosely-coupled distributed service." *Dissertation Abstracts International Part B: Science and Engineering,*, vol. 46, no. 1, 1985.

[9] M. Di Marco, A. Garulli, A. Giannitrapani, and A. Vicino, "SLAM for a team of cooperating robots: a set membership approach," *IEEE Trans. on Robotics and Automation*, vol. 19, no. 2, pp. 238–249, 2003.

[10] A. Fagiolini, M. Pellinacci, G. Valenti, G. Dini, and A. Bicchi, "Consensus–based Distributed Intrusion Detection for Secure Multi-Robot Systems," *IEEE Int. Conf. on Robotics and Automation*, 2008.

[11] F. Robert, "Itérations sur des ensembles finis convergence d'automates cellulaires contractants," *Linear Algebra and its applications*, vol. 29, pp. 393–412, 1980.

[12] ——, "Dérivée discrète et comportement local d'une itération discrète," *Linear algebra and its applications*, vol. 52, pp. 547–589, 1983.

[13] M. Shih and J. Ho, "Solution of the Boolean Markus-Yamabe Problem," *Advances in Applied Mathematics*, vol. 22, no. 1, 1999.

[14] F. Pasqualetti, A. Bicchi, and F. Bullo, "Distributed intrusion detection for secure consensus computations," *IEEE Int. Conf. on Decision and Control*, pp. 5594–5599, 2007.